%% file: main.tex
\documentclass{article}

\usepackage[numbers]{natbib}
\usepackage[preprint]{neurips_2022}

\usepackage[dvipsnames]{xcolor}         
\definecolor{linkColor}{rgb}{0.18,0.39,0.62}
\usepackage[utf8]{inputenc} 
\usepackage[T1]{fontenc}    
\usepackage[colorlinks=true,linkcolor=linkColor,citecolor=linkColor,filecolor=linkColor,urlcolor=linkColor]{hyperref}       
\usepackage{url}            
\usepackage{booktabs}       
\usepackage{amsfonts}       
\usepackage{nicefrac}       
\usepackage{microtype}      

\usepackage{graphicx}
\usepackage{arydshln}
\usepackage{booktabs}
\usepackage{multirow}
\usepackage{caption}
\usepackage{subcaption}
\usepackage{makecell}
\usepackage{csquotes}
\usepackage{epigraph}
\usepackage{amsmath}
\usepackage{xcolor}  
\usepackage{cleveref}
\usepackage{pifont}
\usepackage{listings}

\RequirePackage{algorithm}
\RequirePackage{algorithmic}

\input{settings.tex}
\input{math_commands.tex}

\newcommand\our{\text{BitNet b1.58}}
\newcommand\llama{\text{LLaMA LLM}}

\title{The Era of 1-bit LLMs: \\ All Large Language Models are in 1.58 Bits}

\vspace{-2.8cm}
\author{
Shuming Ma\thanks{~Equal contribution. $\diamond$ Corresponding author. S. Ma, L. Ma, L. Wang, W. Wang, S. Huang, L. Dong, J. Xue, F. Wei are with Microsoft Research. H. Wang and R. Wang are with University of Chinese Academy of Sciences.}~~~~Hongyu Wang\footnotemark[1]~~~~Lingxiao Ma~~~~Lei Wang~~~~Wenhui Wang \\
\bf ~~~~Shaohan Huang~~~~Li Dong~~~~Ruiping Wang~~~~Jilong Xue~~~~Furu Wei$^{\diamond}$ \\
{\href{https://aka.ms/GeneralAI}{https://aka.ms/GeneralAI}}
\vspace{-0.4cm}
\\}

\begin{document}
\maketitle
\vspace{-0.4cm}
\begin{abstract}
\vspace{-0.25cm}
Recent research, such as BitNet~\cite{bitnet}, is paving the way for a new era of 1-bit Large Language Models (LLMs). In this work, we introduce a 1-bit LLM variant, namely \textbf{\our{}}, in which every single parameter (or weight) of the LLM is ternary \{-1, 0, 1\}. It matches the full-precision (i.e., FP16 or BF16) Transformer LLM with the same model size and training tokens in terms of both perplexity and end-task performance, while being significantly more cost-effective in terms of latency, memory, throughput, and energy consumption.
More profoundly, the 1.58-bit LLM defines a new scaling law and recipe for training new generations of LLMs that are both high-performance and cost-effective. Furthermore, it enables a new computation paradigm and opens the door for designing specific hardware optimized for 1-bit LLMs.

\end{abstract}
\vspace{-0.45cm}
\begin{figure}[h]
\centering
\begin{subfigure}{0.8\textwidth}
\centering
\includegraphics[width=\linewidth]{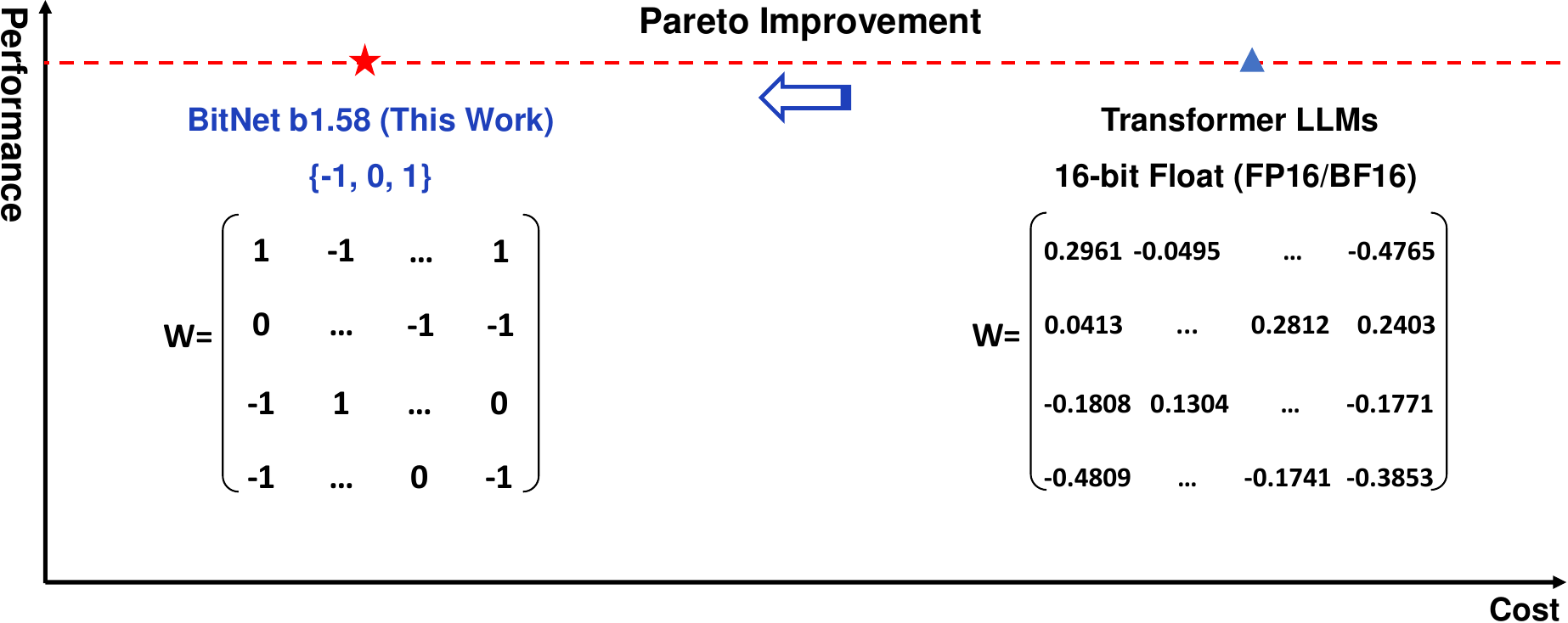}
\vspace{-6pt}
\end{subfigure}
\begin{subfigure}{0.8\textwidth}
\centering
\includegraphics[width=\linewidth]{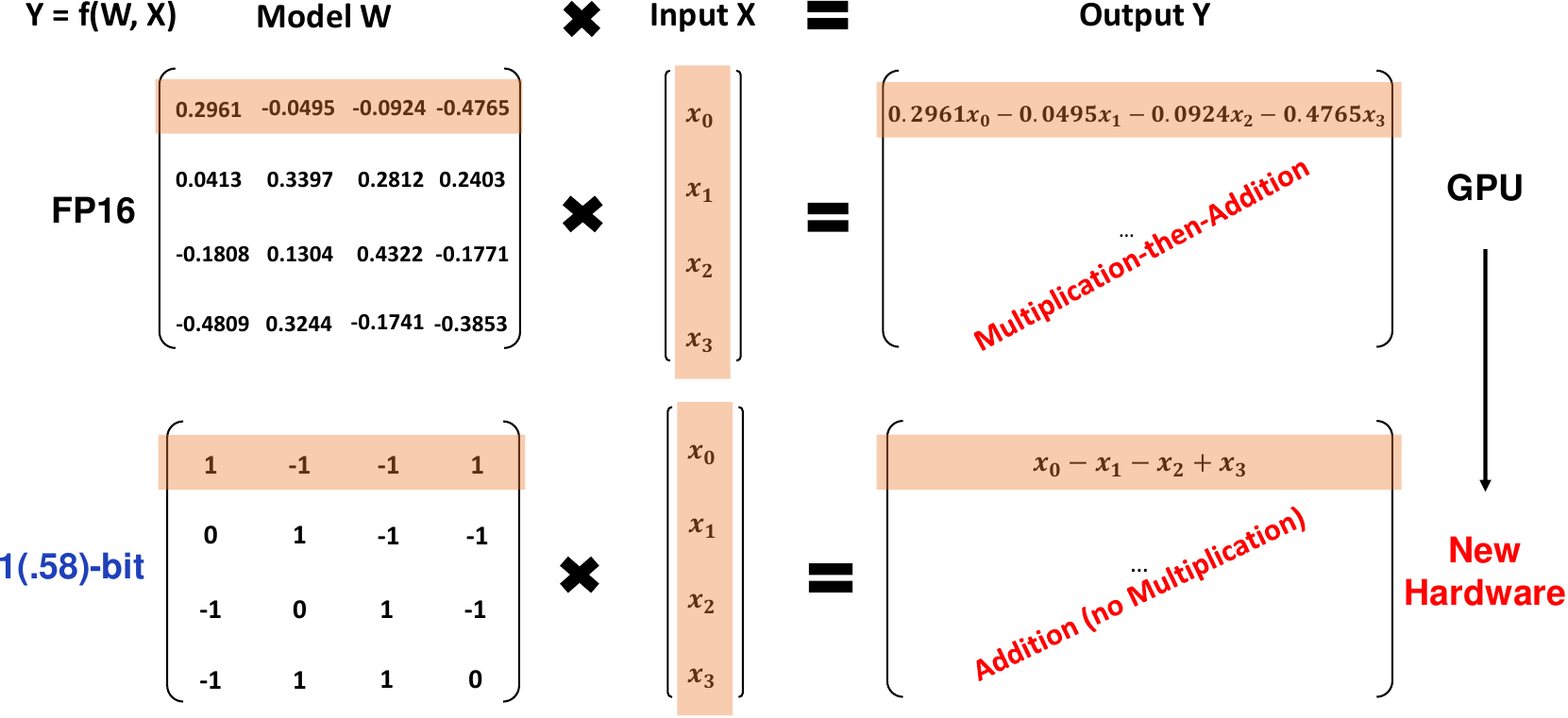}
\end{subfigure}
\caption{1-bit LLMs (e.g., \our{}) provide a Pareto solution to reduce inference cost (latency, throughput, and energy) of LLMs while maintaining model performance. The new computation paradigm of \our{} calls for actions to design new hardware optimized for 1-bit LLMs.}
\end{figure}


\section{The Era of 1-bit LLMs}

In recent years, the field of AI has seen a rapid growth in the size and capabilities of Large Language Models (LLMs). These models have demonstrated remarkable performance in a wide range of natural language processing tasks, but their increasing size has posed challenges for deployment and raised concerns about their environmental and economic impact due to high energy consumption. 
One approach to address these challenges is to use post-training quantization to create low-bit models for inference~\cite{smoothquant, gptq, quip, quip_sharp}. This technique reduces the precision of weights and activations, significantly reducing the memory and computational requirements of LLMs. The trend has been to move from 16 bits to lower bits, such as 4-bit variants~\cite{gptq, awq}. However, post-training quantization is sub-optimal, even though it is widely used in industry LLMs.

Recent work on 1-bit model architectures, such as BitNet~\cite{bitnet}, presents a promising direction for reducing the cost of LLMs while maintaining their performance. Vanilla LLMs are in 16-bit floating values (i.e., FP16 or BF16), and the bulk of any LLMs is matrix multiplication. Therefore, the major computation cost comes from the floating-point addition and multiplication operations. In contrast, the matrix multiplication of BitNet only involves integer addition, which saves orders of energy cost for LLMs. As the fundamental limit to compute performance in many chips is power, the energy savings can also be translated into faster computation.

In addition to computation, the process of transferring model parameters from DRAM to the memory of an on-chip accelerator (e.g., SRAM) can be expensive during inference. There have been attempts to enlarge SRAM to improve throughput, but this introduces significantly higher costs than DRAM. Compared to full-precision models, 1-bit LLMs have a much lower memory footprint from both a capacity and bandwidth standpoint. This can significantly reduce the cost and time of loading weights from DRAM, leading to faster and more efficient inference.

In this work, we introduce a significant 1-bit LLM variant called \textbf{\our{}}, where every parameter is ternary, taking on values of \{-1, 0, 1\}. We have added an additional value of 0 to the original 1-bit BitNet, resulting in 1.58 bits in the binary system. \our{} retains all the benefits of the original 1-bit BitNet, including its new computation paradigm, which requires almost no multiplication operations for matrix multiplication and can be highly optimized. Additionally, it has the same energy consumption as the original 1-bit BitNet and is much more efficient in terms of memory consumption, throughput and latency compared to FP16 LLM baselines. Furthermore, \our{} offers two additional advantages. Firstly, its modeling capability is stronger due to its explicit support for feature filtering, made possible by the inclusion of 0 in the model weights, which can significantly improve the performance of 1-bit LLMs. Secondly, our experiments show that \our{} can match full precision (i.e., FP16) baselines in terms of both perplexity and end-task performance, starting from a 3B size, when using the same configuration (e.g., model size, training tokens, etc.).

\section{BitNet b1.58}

\our{} is based on the BitNet architecture, which is a Transformer that replaces \emph{nn.Linear} with \emph{BitLinear}. It is trained from scratch, with 1.58-bit weights and 8-bit activations. Compared to the original BitNet, it introduces some modifications that we summarize below.

\paragraph{Quantization Function.} To constrain the weights to -1, 0, or +1, we adopt an \emph{absmean} quantization function. It first scales the weight matrix by its average absolute value, and then round each value to the nearest integer among \{-1, 0, +1\}:

\begin{equation}
    \widetilde{W} = \mathrm{RoundClip}(\frac{W}{\gamma + \epsilon}, -1, 1),
\end{equation}
\begin{equation}
    \mathrm{RoundClip}(x, a, b) = \max(a, \min(b, \mathrm{round}(x))),
\end{equation}
\begin{equation}
    \gamma = \frac{1}{nm}\sum_{ij} |W_{ij}|.
\end{equation}

The quantization function for activations follows the same implementation in BitNet, except that we do not scale the activations before the non-linear functions to the range $[0, Q_b]$. Instead, the activations are all scaled to $[-Q_b, Q_b]$ per token to get rid of the zero-point quantization. This is more convenient and simple for both implementation and system-level optimization, while introduces negligible effects to the performance in our experiments.

\paragraph{LLaMA-alike Components.}

The architecture of LLaMA~\cite{llama, llama2} has been the de-facto backbone for open-source LLMs. To embrace the open-source community, our design of \our{} adopts the LLaMA-alike components. Specifically, it uses RMSNorm~\cite{rmsnorm}, SwiGLU~\cite{swiglu}, rotary embedding~\cite{rope}, and removes all biases. In this way, \our{} can be integrated into the popular open-source software (e.g., Huggingface, vLLM~\cite{vllm}, and llama.cpp\footnote{\url{https://github.com/ggerganov/llama.cpp}}) with minimal efforts.

\begin{table*}[t]
\setlength{\tabcolsep}{10pt}
\centering
\begin{tabular}{lcccc}
\toprule
\textbf{Models} & \textbf{Size} & \textbf{Memory (GB)$\downarrow$} & \textbf{Latency (ms)$\downarrow$} & \textbf{PPL$\downarrow$} \\
\midrule
\llama & 700M & 2.08 (1.00x) & 1.18 (1.00x)  & 12.33 \\
\textbf{\our{}} & 700M & 0.80 (2.60x) & 0.96 (1.23x) & 12.87 \\
\midrule
\llama & 1.3B & 3.34 (1.00x) & 1.62 (1.00x) & 11.25 \\
\textbf{\our{}} & 1.3B & 1.14 (2.93x) & 0.97 (1.67x)& 11.29 \\
\midrule
\llama & 3B & 7.89 (1.00x) & 5.07 (1.00x) & 10.04 \\
\textbf{\our{}} & 3B & \textbf{2.22 (3.55x)} & \textbf{1.87 (2.71x)} & \textbf{9.91} \\
\textbf{\our{}} & 3.9B & \textbf{2.38 (3.32x)} & \textbf{2.11 (2.40x)} & \textbf{9.62} \\
\bottomrule
\end{tabular}
\caption{Perplexity as well as the cost of \our{} and \llama{}.}
\label{tab:ppl}
\end{table*}

\begin{table*}[t]
\setlength{\tabcolsep}{6pt}
\centering
\begin{tabular}{lccccccccc}
\toprule
\textbf{Models} & \textbf{Size} & \textbf{ARCe} & \textbf{ARCc} & \textbf{HS} & \textbf{BQ} & \textbf{OQ} & \textbf{PQ} & \textbf{WGe} & \textbf{Avg.} \\
\midrule
\llama & 700M & 54.7 & 23.0 & 37.0 & 60.0 & 20.2 & 68.9 & 54.8 & 45.5\\
\textbf{\our{}} & 700M & 51.8 & 21.4 & 35.1 & 58.2 & 20.0 & 68.1 & 55.2 & 44.3\\
\midrule
\llama & 1.3B & 56.9 & 23.5 & 38.5 & 59.1 & 21.6 & 70.0 & 53.9 & 46.2\\
\textbf{\our{}} & 1.3B & 54.9 & 24.2 & 37.7 & 56.7 & 19.6 & 68.8 & 55.8 & 45.4\\
\midrule
\llama & 3B & 62.1 & 25.6 & 43.3 & 61.8 & 24.6 & 72.1& 58.2 & 49.7\\
\textbf{\our{}} & 3B & \textbf{61.4} & \textbf{28.3} & \textbf{42.9} & \textbf{61.5} & \textbf{26.6} & \textbf{71.5} & \textbf{59.3} & \textbf{50.2}\\
\textbf{\our{}} & 3.9B & \textbf{64.2} & \textbf{28.7} & \textbf{44.2} & \textbf{63.5} & \textbf{24.2} & \textbf{73.2} & \textbf{60.5} & \textbf{51.2} \\
\bottomrule
\end{tabular}
\caption{Zero-shot accuracy of \our{} and \llama{} on the end tasks.}
\label{tab:zero-shot}
\end{table*}

\section{Results}

We compared \our{} to our reproduced FP16 \llama{} in various sizes. To ensure a fair comparison, we pre-trained the models on the RedPajama dataset~\cite{redpajama} for 100 billion tokens. We evaluated the zero-shot performance on a range of language tasks, including ARC-Easy~\cite{arc}, ARC-Challenge~\cite{arc}, Hellaswag~\cite{hellaswag}, Winogrande~\cite{winoGrande}, PIQA~\cite{piqa}, OpenbookQA~\cite{openbookqa}, and BoolQ~\cite{boolq}. We also reported the validation perplexity on the WikiText2~\cite{wikitext2} and C4~\cite{c4} datasets.

We compared the runtime GPU memory and latency of both \llama{} and \our{}. The results were measured using the FasterTransformer\footnote{\url{https://github.com/NVIDIA/FasterTransformer}} codebase, which is well-optimized for LLM inference latency on GPU devices. The 2-bit kernel from Ladder~\cite{ladder} is also integrated for \our{}. We reported the time per output token, as it is the major cost for inference.

Table~\ref{tab:ppl} summarizes the perplexity and the cost for \our{} and \llama{}. It shows that \our{} starts to match full precision \llama{} at 3B model size in terms of perplexity, while being 2.71 times faster and using 3.55 times less GPU memory. In particular, \our{} with a 3.9B model size is 2.4 times faster, consumes 3.32 times less memory, but performs significantly better than \llama{} 3B. 

Table~\ref{tab:zero-shot} reports the detailed results of the zero-shot accuracy on the end tasks. We followed the pipeline from \emph{lm-evaluation-harness}\footnote{\url{https://github.com/EleutherAI/lm-evaluation-harness}} to perform the evaluation. The results show that the performance gap between \our{} and \llama{} narrows as the model size increases. More importantly, \our{} can match the performance of the full precision baseline starting from a 3B size. Similar to the observation of the perplexity, the end-task results reveal that \our{} 3.9B outperforms \llama{} 3B with lower memory and latency cost. This demonstrates that \our{} is a Pareto improvement over the state-of-the-art LLM models.

\paragraph{Memory and Latency}

\begin{figure}[t]
    \centering
    \includegraphics[width=0.49\textwidth]{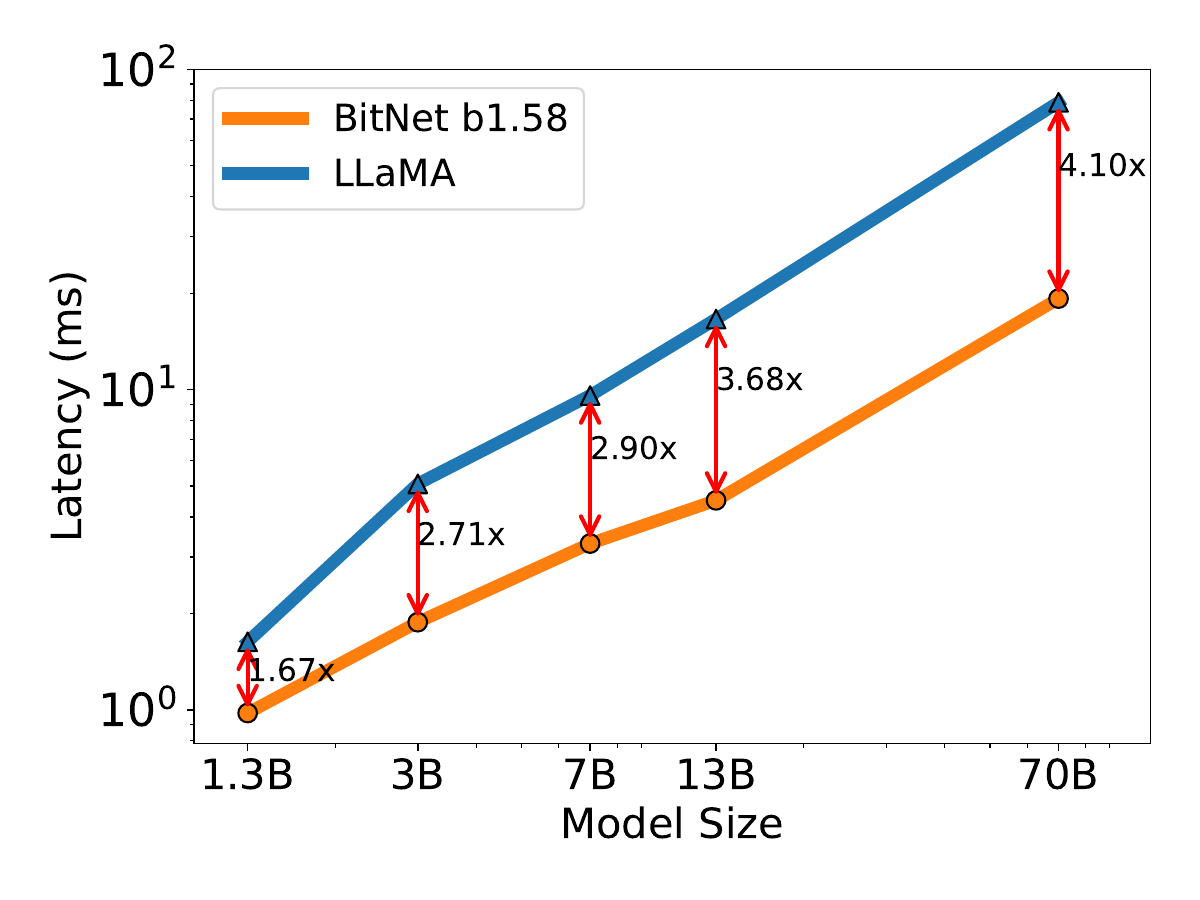}
    \includegraphics[width=0.49\textwidth]{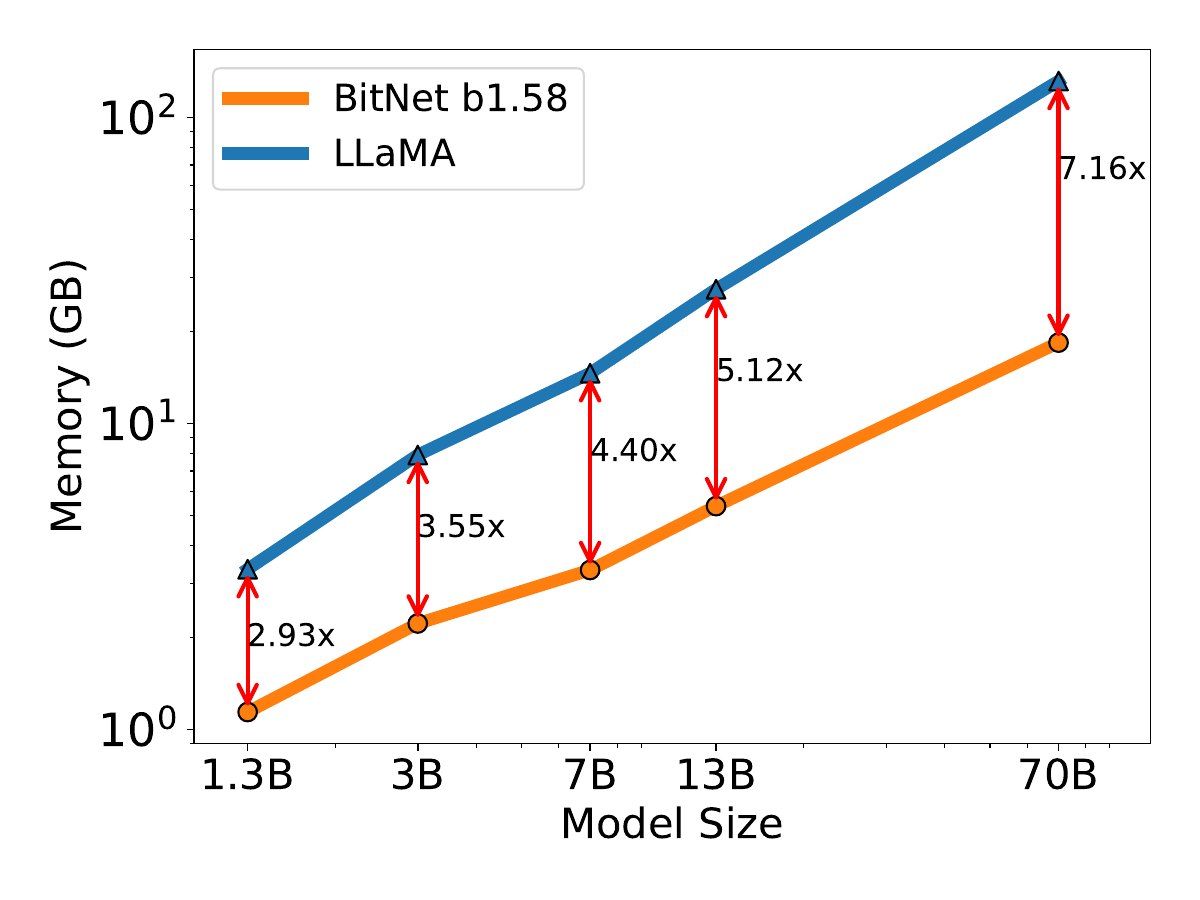}
    \caption{Decoding latency (Left) and memory consumption (Right) of \our{} varying the model size.}
    \label{fig:latency-memory-energy}
\end{figure}

We further scaled up the model size to 7B, 13B, and 70B and evaluated the cost. Figure~\ref{fig:latency-memory-energy} illustrates the trends of latency and memory, showing that the speed-up increases as the model size scales. In particular, \our{} 70B is 4.1 times faster than the \llama{} baseline. This is because the time cost for \emph{nn.Linear} grows with the model size. The memory consumption follows a similar trend, as the embedding remains full precision and its memory proportion is smaller for larger models. Both latency and memory were measured with a 2-bit kernel, so there is still room for optimization to further reduce the cost.

\paragraph{Energy}

\begin{figure}[t]
    \centering
    \includegraphics[width=0.49\textwidth]{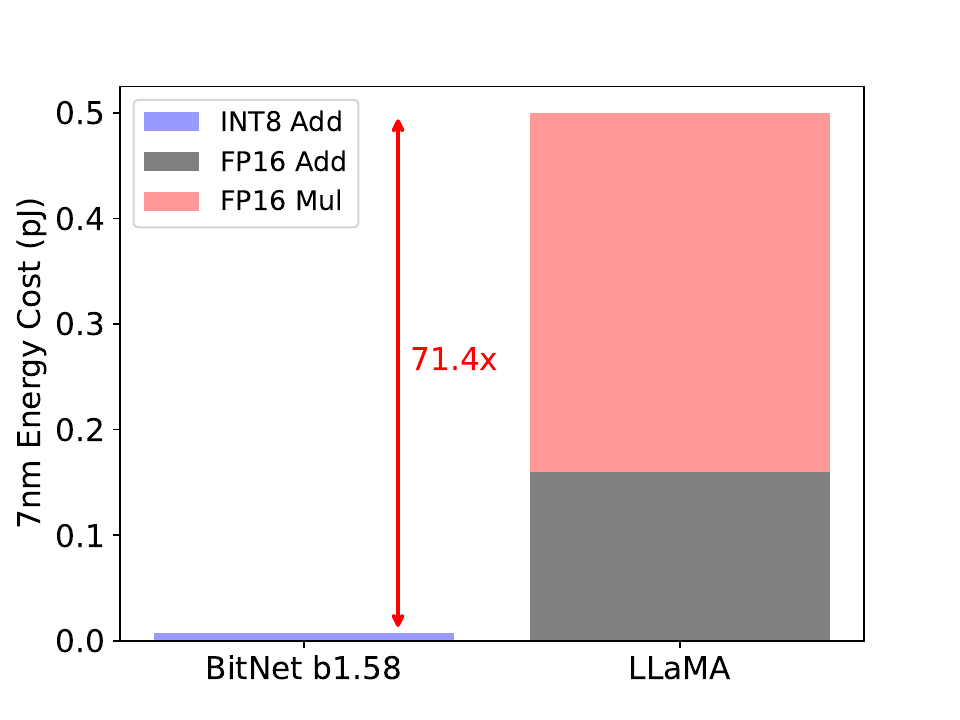}
    \includegraphics[width=0.49\textwidth]{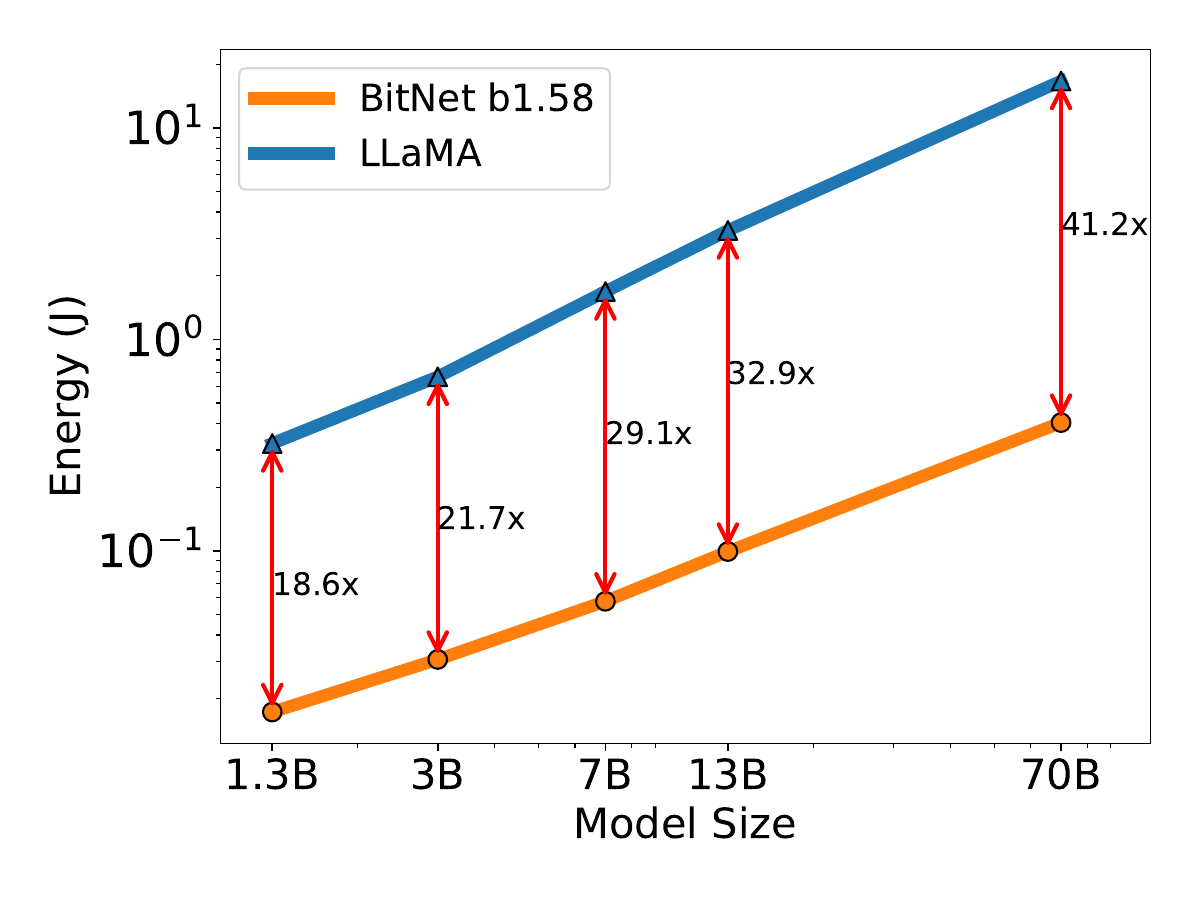}
    \caption{Energy consumption of \our{} compared to \llama{} at 7nm process nodes. On the left is the components of arithmetic operations energy. On the right is the end-to-end energy cost across different model sizes.}
    \label{fig:energy}
\end{figure}

We also estimate the arithmetic operations energy consumption of both \our{} and \llama{}.
We focus mainly on the calculation for matrix multiplication, since it
contributes the most to the cost of LLMs. Figure~\ref{fig:energy} illustrates the composition of the energy cost. The majority of \our{} is INT8 addition calculation, while \llama{} consists of both FP16 addition and FP16 multiplication. According to the energy model in~\cite{energycost, pokebnn}, \our{} saves 71.4 times arithmetic operations energy consumption for matrix multiplication on 7nm chips. We further reported the end-to-end energy cost for models with 512 tokens. Our results show that as the model size scales, \our{} becomes increasingly more efficient in terms of energy consumption compared to the FP16 \llama{} baseline. This is due to the fact that the percentage of \emph{nn.Linear} grows with the model size, while the cost from other components is smaller for larger models.

\paragraph{Throughput}

\begin{table*}[t]
\setlength{\tabcolsep}{10pt}
\centering
\begin{tabular}{lccc}
\toprule
\textbf{Models} & \textbf{Size} & \textbf{Max Batch Size} & \textbf{Throughput (tokens/s)} \\
\midrule
\llama{} & 70B & 16 (1.0x) & \ \ 333 (1.0x) \\
\textbf{\our{}} & 70B & \textbf{176 (11.0x)} & \textbf{2977 (8.9x)}\\
\bottomrule
\end{tabular}
\caption{Comparison of the throughput between \our{} 70B and \llama{} 70B.}
\label{tab:throughput}
\end{table*}

We compare the throughput of \our{} and \llama{} with 70B parameters on two 80GB A100 cards, using pipeline parallelism~\cite{gpipe} so that \llama{} 70B could be run on the devices. We increased the batch size until the GPU memory limit was reached, with a sequence length of 512. Table~\ref{tab:throughput} shows that \our{} 70B can support up to 11 times the batch size of \llama{}, resulting an 8.9 times higher throughput.

\textbf{\our{} is enabling a new scaling law with respect to model performance and inference cost}. As a reference, we can have the following equivalence between different model sizes in 1.58-bit and 16-bit based on the results in Figure~\ref{fig:latency-memory-energy} and \ref{fig:energy}.
\begin{itemize}
    \item 13B BitNet b1.58 is more efficient, in terms of latency, memory usage and energy consumption, than 3B FP16 LLM.
    \item 30B BitNet b1.58 is more efficient, in terms of latency, memory usage and energy consumption, than 7B FP16 LLM.
    \item 70B BitNet b1.58 is more efficient, in terms of latency, memory usage and energy consumption, than 13B FP16 LLM.
\end{itemize}

\paragraph{Training with 2T Tokens}

The number of training tokens is a crucial factor for LLMs. To test the scalability of \our{} in terms of tokens, we trained a \our{} model with 2T tokens following the data recipe of StableLM-3B~\cite{StableLM-3B-4E1T}, which is the state-of-the-art open-source 3B model. Both models were evaluated on a benchmark that consists of Winogrande~\cite{winoGrande}, PIQA~\cite{piqa}, SciQ~\cite{sciq}, LAMBADA~\cite{lambada}, and ARC-easy~\cite{arc}. We reported the zero-shot accuracy in Table~\ref{tab:2t}. For tasks measured with accuracy and normalized accuracy, we take the average of the two. The results of StableLM 3b at 2T tokens are taken directly from its technical report. Our findings shows that \our{} achieves a superior performance on all end tasks, indicating that 1.58-bit LLMs also have strong generalization capabilities.

\begin{table*}[t]
\resizebox{\linewidth}{!}{
\setlength{\tabcolsep}{6pt}
\centering
\begin{tabular}{lcccccccc}
\toprule
\textbf{Models} & \textbf{Tokens} & \textbf{Winogrande} & \textbf{PIQA} & \textbf{SciQ} & \textbf{LAMBADA} & \textbf{ARC-easy} & \textbf{Avg.} \\
\midrule
StableLM-3B & 2T & 64.56 & 76.93 & 90.75 & 66.09 & 67.78 & 73.22 \\
\textbf{\our{}} 3B & 2T & \textbf{66.37} & \textbf{78.40} & \textbf{91.20} & \textbf{67.63} & \textbf{68.12} & \textbf{74.34} \\
\bottomrule
\end{tabular}
}
\caption{Comparison of \our{} with StableLM-3B with 2T tokens.}
\vspace{15pt}
\label{tab:2t}
\end{table*}

\section{Discussion and Future Work}

\textbf{1-bit Mixture-of-Experts (MoE) LLMs}

Mixture-of-Experts (MoE) have proven to be a cost-effective approach for LLMs. While it significantly reduces the computation FLOPs, the high memory consumption and inter-chip communication overhead limit its deployment and application. These challenges can be addressed by 1.58-bit LLMs. Firstly, the reduced memory footprint reduces the number of devices required to deploy MoE models. Moreover, it significantly reduces the overhead of transferring activations across networks. Ultimately, there would be no overhead if the entire models could be placed on a single chip.

\textbf{Native Support of Long Sequence in LLMs}

In the era of LLMs, the ability to handle long sequence has become a critical demand. One major challenge for long sequence inference is the memory consumption introduced by the KV caches. \our{} represents a significant step towards native support for long sequences, as it reduces the activations from 16 bits to 8 bits, allowing the context length to be doubled given the same resources. 
This can be further losslessly compressed to 4 bits or even lower for 1.58-bit LLMs, which we leave as future work.

\textbf{LLMs on Edge and Mobile}

The use of 1.58-bit LLMs has the potential to greatly improve the performance of language models on edge and mobile devices. These devices are often limited by their memory and computational power, which can restrict the performance and the scale of LLMs. However, the reduced memory and energy consumption of 1.58-bit LLMs allows them to be deployed on these devices, enabling a wide range of applications that were previously not possible. This can greatly enhance the capabilities of edge and mobile devices and enable new and exciting applications of LLMs. Moreover, 1.58-bit LLMs are more friendly to CPU devices, which are the main processors used in edge and mobile devices. This means that \our{} can be efficiently executed on these devices, further improving their performance and capabilities.

\textbf{New Hardware for 1-bit LLMs}

Recent work like~Groq\footnote{\url{https://groq.com/}} has demonstrated promising results and great potential for building specific hardware (e.g., LPUs) for LLMs. Going one step further, we envision and call for actions to design new hardware and system specifically optimized for 1-bit LLMs, given the new computation paradigm enabled in BitNet~\cite{bitnet}.

\bibliography{bitnet}
\bibliographystyle{alpha}

\end{document}

%% file: settings.tex
\usepackage{multirow}
\usepackage{amsmath}
\usepackage{capt-of}
\usepackage{tabularx}
\usepackage{epsfig}
\usepackage{amssymb}
\usepackage{amsfonts}
\usepackage{booktabs}
\usepackage{scalerel}
\usepackage[inline]{enumitem}
\usepackage{listings}
\usepackage{varwidth}
\usepackage[export]{adjustbox}
\usepackage{tikz}
\usetikzlibrary{tikzmark}

\usepackage{stmaryrd}
\usepackage{bbm}
\usepackage{wrapfig}
\usepackage{pifont}

\definecolor{deepblue}{rgb}{0,0,0.5}
\definecolor{officeblue}{RGB}{0,102,204}
\definecolor{deepred}{rgb}{0.6,0,0}
\definecolor{deepgreen}{rgb}{0,0.5,0}
\definecolor{mybrickred}{RGB}{182,50,28}

\definecolor{fillcolor}{RGB}{216,217,252}

\usepackage{etoolbox}
\usepackage{framed}

\newif\ifxetexorluatex
\ifxetex
  \xetexorluatextrue
\else
  \ifluatex
    \xetexorluatextrue
  \else
    \xetexorluatexfalse
  \fi
\fi
\newcommand*\quotesize{60} 
\newcommand*{\openquote}
   {\tikz[remember picture,overlay,xshift=-4ex,yshift=-2.5ex]
   \node (OQ) {\fontsize{\quotesize}{\quotesize}\selectfont``};\kern0pt}

\newcommand*{\closequote}[1]
  {\tikz[remember picture,overlay,xshift=4ex,yshift={#1}]
   \node (CQ) {\fontsize{\quotesize}{\quotesize}\selectfont''};}

\colorlet{shadecolor}{white}

\newcommand*\shadedauthorformat{\emph} 

\newcommand*\authoralign[1]{%
  \if#1l
    \def\authorfill{}\def\quotefill{\hfill}
  \else
    \if#1r
      \def\authorfill{\hfill}\def\quotefill{}
    \else
      \if#1c
        \gdef\authorfill{\hfill}\def\quotefill{\hfill}
      \else\typeout{Invalid option}
      \fi
    \fi
  \fi}
{\authoralign{#1}
\ifblank{#2}
   {\def\shadequoteauthor{}\def\yshift{-2ex}\def\quotefill{\hfill}}
   {\def\shadequoteauthor{\par\authorfill\shadedauthorformat{#2}}\def\yshift{2ex}}
\begin{snugshade}\begin{quote}\openquote}
{\shadequoteauthor\quotefill\closequote{\yshift}\end{quote}\end{snugshade}}

%% file: math_commands.tex
\usepackage{amsmath,amsfonts,bm}

\def\eqref#1{equation~\ref{#1}}

\def\1{\bm{1}}

\DeclareMathAlphabet{\mathsfit}{\encodingdefault}{\sfdefault}{m}{sl}
\SetMathAlphabet{\mathsfit}{bold}{\encodingdefault}{\sfdefault}{bx}{n}

